# A Novel Clustering Algorithm Based on Quantum Random Walk

Qiang Li, Yan He, Jing-ping Jiang

College of Electrical Engineering, Zhejiang University,
Hang Zhou, Zhejiang, 310027, China

October 23, 2018

**Abstract**

The enormous successes have been made by quantum algorithms during the last decade. In this paper, we combine the quantum random walk (QRW) with the problem of data clustering, and develop two clustering algorithms based on the one dimensional QRW. Then, the probability distributions on the positions induced by QRW in these algorithms are investigated, which also indicates the possibility of obtaining better results. Consequently, the experimental results have demonstrated that data points in datasets are clustered reasonably and efficiently, and the clustering algorithms are of fast rates of convergence. Moreover, the comparison with other algorithms also provides an indication of the effectiveness of the proposed approach.

**Keywords**: Unsupervised learning; Data clustering; Quantum computation; Quantum random walk

## 1 Introduction

Quantum computation is an extremely exciting and rapidly growing field of investigation, and has attracted a lot of interests. More recently, an increasing number of researchers with different backgrounds, ranging from physics, computer sciences and information theory to mathematics and philosophy, are involved in researching properties of quantum-based computation [1]. During the last decade, a series of significant breakthroughs have been made. One was that in 1994 Peter Shor surprised the world by describing a polynomial time quantum algorithm for factoring integers [2], while in classical world this was a NP-complete problem that didn't find an efficient algorithm. Three years later, in 1997, Lov Grover proved that a quantum computer could search an unsorted database in only the square root of the time [3]. Meanwhile, Gilles Brassard et al. combined ideas from Grover's and Shor's quantum algorithms to propose a quantum counting algorithm [4].

In recent years, many interests focus on quantum random walks (QRW) and considerable work has been done [5, 6, 7]. For instance, some bounds were given on general graphs [8], where, for the standard deviation and the mixing time, a quadratic speed up were reported over the classical counterparts on the line and cycle. Later, J. Kempe [9] proved that the hitting time from one corner to the opposite corner on a $n$-bit hypercube showed an exponential speed up, and A. M. Childs et al. [10] used a continuous time quantum walk to traverse a special graph exponentially faster than any classical algorithms.

Successes that have been made by quantum algorithms make us guess that powerful quantum computers can figure out solutions faster and better than the best known classical counterparts, or even solve certain problems that classical computer cannot solve. Furthermore, it is more important that they offer a new way to find potentially dramatic algorithmic speed-ups. Therefore, we may ask naturally: can we construct quantum versions of classical algorithms or present new quantum algorithms to solve the problems in pattern recognition faster and better on the quantum computer? Following this idea, some pioneers have proposed their novel methods and demonstrated exciting consequences [11, 12, 13].

In this paper, we attempt to combine the QRW with the problem of data clustering in order to establish a novel QRW based clustering algorithm. QRWs differ from classical random walks in that their evolution is unitary and reversible. In the discrete case, an extra "coin" degree of freedom (usually a single quantum bit) is introduced into the system. Just like the classical random walk, the particle's moves depend on the outcome of a "coin flip". However, in the quantum case, both the "coin flip" and the conditional shift of the particle are unitary transformations, and different possible classical paths can interfere with each other.

In our algorithms, data points in a dataset are viewed as particles that can walk at random in an $m$-dimensional metric space according to certain rules. Further, each data point may be regarded as a local control subsystem, whose controller controls its walking behavior. From the point of view of control theory, the system with $N$ particles which walk randomly in space may be described by the below block diagram.

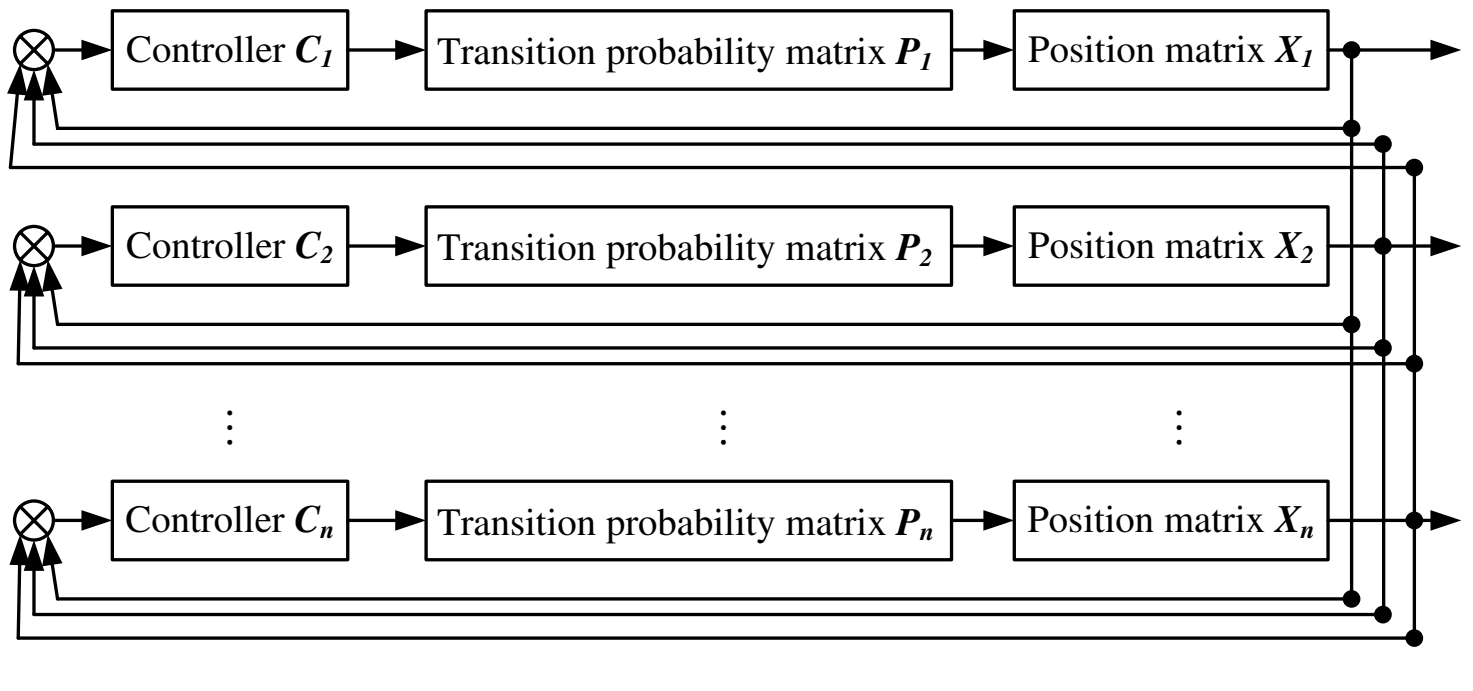

Figure 1: The block diagram of the system.

As is shown in Fig. 1, the controlled object is the Transition Probability matrix $P$, and the outputs of the system are the new positions of all particles in the system. The controller $C$ adjusts the entries in the Transition Probability matrix $P$ according to the current positions of $N$ particles, and then decides the transition directions and transition distances of $N$ particles at the next moment. Finally, the positions of $N$ particles are updated synchronously. As data points move in the space at random, they gather together gradually and form some separating clusters automatically.

The remainder of this paper is organized as follows: Section 2 introduces some important concepts about the quantum computation and the quantum random walk briefly. Section 3 elaborates two proposed clustering algorithms based on one dimensional QRW. Section 4 discusses the relationship between the number of clusters and the number of nearest neighbors firstly, and then the effects of number of steps in the 1D-scms and 1D-mcms algorithms are investigated. Section 5 introduces those datasets used in the experiments briefly, and then compares experimental results of the proposed algorithms with other clustering algorithms. The conclusion is given in Section6.

# 2 Quantum computation and Quantum random walk

## 2.1 Quantum computation [1, 14]

The elementary unit of quantum computation is called the qubit, which is typically a microscopic system, such as an atom, a nuclear spin, or a polarized photon. In quantum computation, the Boolean states 0 and 1 are represented by a prescribed pair of normalized and mutually orthogonal quantum states labeled as $\{|0\rangle, |1\rangle\}$ to form a 'computational basis'. Any pure state of the qubit can be written as a superposition state $\alpha|0\rangle + \beta|1\rangle$ for some $\alpha$ and $\beta$ satisfying $|\alpha|^2 + |\beta|^2 = 1$. A collection of $n$ qubits is called a quantum register of size $n$, which spans a Hilbert space of $2^n$ dimensions, and so $2^n$ mutually orthogonal quantum states can be available.

Quantum state preparations, and any other manipulations on qubits, have to be performed by unitary operations. A quantum logic gate is a device which performs a fixed unitary operation on selected qubits in a fixed period of time, and a quantum circuit is a device consisting of quantum logic gates whose computational steps are synchronized in time . The most common quantum gate is the Hadamard gate, which acts on a qubit in state $|0\rangle$ or $|1\rangle$ to produce

$$H = \frac{1}{\sqrt{2}} \begin{pmatrix} 1 & 1 \\ 1 & -1 \end{pmatrix}, \begin{cases} |0\rangle \xrightarrow{H} \frac{1}{\sqrt{2}}|0\rangle + \frac{1}{\sqrt{2}}|1\rangle \\ |1\rangle \xrightarrow{H} \frac{1}{\sqrt{2}}|0\rangle - \frac{1}{\sqrt{2}}|1\rangle. \end{cases} \tag{1}$$

## 2.2 Quantum random walk [7, 15]

In this subsection, we focus on the discrete model of the quantum random walk in one dimension whose notion will be formally defined. First, let $H_P$ be the Hilbert space spanned by basis states $\{|i\rangle\}$ representing the position of the particles, and $H_C$ be the two-dimensional coin space spanned by two basis states $\{|\Uparrow\rangle, |\Downarrow\rangle\}$. So the total Hilbert space is given by $H = H_P \otimes H_C$. Just like the classical random walk, the evolution of a quantum random walk is divided into two subsequent operations: the coin operation and the conditional shift operation.

The coin operation $C$ rotates a state, $|\uparrow\rangle$ or $|\downarrow\rangle$, in $H_C$ to render the coin state in a superposition, which is in analogy to the coin flip in classical random walk. One can design different unitary transformation $C$ to observe different behavior of walks, while the Hadamard coin, a balanced coin, is commonly used, which gives equal chances to move left and right.

For the conditional shift operation $S$, it makes the walker take a step to the right or to the left in terms of the accompanying coin state, which has the form

$$S = |\uparrow\rangle\langle\uparrow| \otimes \sum_i |i+1\rangle\langle i| + |\downarrow\rangle\langle\downarrow| \otimes \sum_i |i-1\rangle\langle i|. \tag{2}$$

Therefore, the evolution of the system at each step of the walk can be described by the total unitary transformation $U = S \cdot (C \otimes I)$, where $I$ is the identity operator on $H_P$. If the transformation $U$ is applied to an initial state $T > 2$ times, a different probability distribution will be yielded before the state are measured. For example, assume the initial state is $|\psi_0\rangle = |\uparrow\rangle \otimes |0\rangle$, after $T = 100$ steps, the probability distribution on the positions of the particle is shown in Fig. 2, where the other curve represents the probability distribution of a classical random walk.

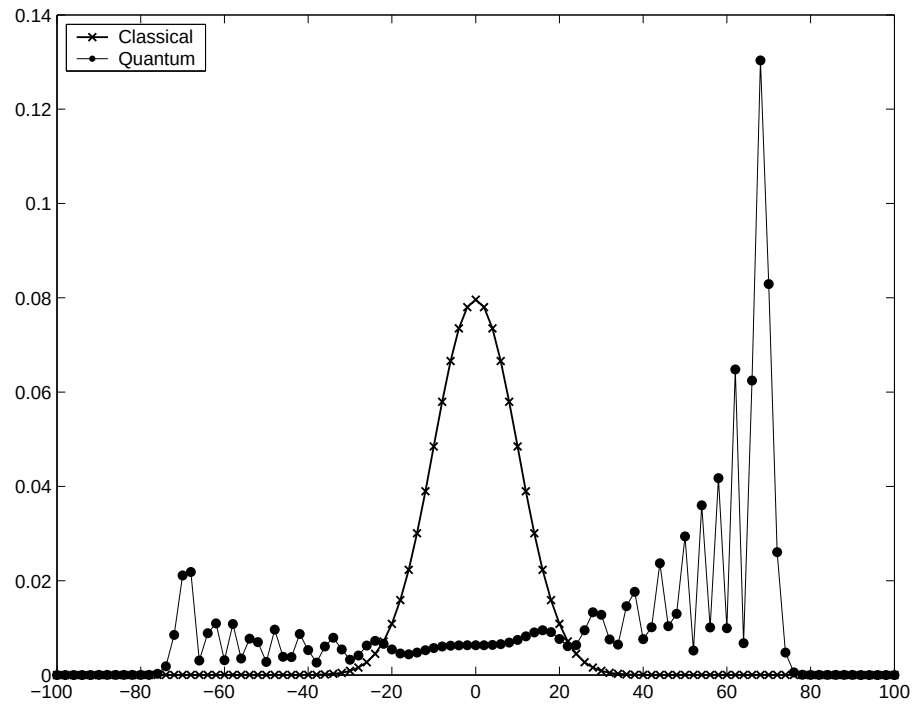

Figure 2: The probability distribution for a classical random walk and a quantum random walk.

# 3 Algorithms

One dimensional QRW as a standard model has been widely investigated [5, 7]. On the basis of this fact, for a problem of data clustering using QRW in a high dimensional space, a natural idea is to divide the $m$-dimensional QRW into $m$ one-dimensional-QRWs. In addition, assume an unlabeled dataset $\boldsymbol{X} = \{\boldsymbol{X}_0^1, \boldsymbol{X}_0^2, \cdots, \boldsymbol{X}_0^n\}$, whose each instance is with $m$ features. In the clustering algorithm based on the QRW, each data point in the dataset is regarded as a movable particle which can walk in the whole space according to transition probabilities. In this section, two clustering algorithms based on one dimensional QRW are constructed which are named respectively: (a) 1D-single-coin-multi-step, (b) 1D-multi-coin-multi-step.

## 3.1 1D-single-coin-multi-step (1D-scms) algorithm

In all clustering algorithms throughout this paper, a decentralized control strategy is employed, i.e., each data point in the dataset only interacts with its neighbors in its neighborhood who are selected by a $k$-nearest-neighbor method. Next, the transition probabilities, $p_t(i,j), j \in \Gamma_t(i)$, are computed as below,

$$
\begin{aligned}
p_t(i,j) &= \begin{cases} \frac{a_t(i,j)}{\sum_{j\in\Gamma_t(i)} a_t(i,j)} & \text{if } j \in \Gamma_t(i) \\ 0 & \text{otherwise} \end{cases} \\
a_t(i,j) &= \frac{\left(Deg_t(j)/\sum_{j\in\Gamma_t(i)} Deg_t(j)\right)\times\left(Deg_0(j)/\sum_{j\in\Gamma_t(i)} Deg_0(j)\right)}{\left(d\left(\boldsymbol{X}_t^i,\boldsymbol{X}_t^j\right)\right)\times\left(d\left(\boldsymbol{X}_0^i,\boldsymbol{X}_0^j\right)\right)}
\end{aligned} \tag{3}
$$

where the variable $\Gamma_t(i)$ represents a neighbor set of a data point $\boldsymbol{X}_t^i$; $Deg_t(\cdot)$ and $Deg_0(\cdot)$ denote the current and initial degrees of a data point respectively; likewise, $d\left(\boldsymbol{X}_t^i,\boldsymbol{X}_t^j\right)$ and $d\left(\boldsymbol{X}_0^i,\boldsymbol{X}_0^j\right)$ denote the current and initial distances of the data point respectively. And then the largest transition probability $p_t(i,h)$ and the neighbor with this probability $\boldsymbol{X}_t^h, h \in \Gamma_t(i), h \neq i$ are identified.

As mentioned above, the quantum random walk in an $m$-dimensional space is divided into $m$ one-dimensional-QRWs. So, for each one of $m$ one-dimensional-walks, a data point $\boldsymbol{X}_t^i$ only can move a step of length $l_L$ or $l_R$ either left or right. Therefore, the maximal transition probability $p_t(i,h)$ needs to be mapped into an interval, $\rho = f(p_t(i,h)) \in [0.5, 1]$, while the probability moving in the opposite direction is $1-\rho$. If $\rho = 0.5$, the Hadamard transformation $H$ may be applied as a balanced unitary coin in the one dimensional quantum random walk. In general case, however, $\rho \neq 1-\rho$, the balanced unitary coin is replaced by a bias coin called the transformation $C$ which is given below,

$$
C = \begin{pmatrix} \sqrt{\rho} & \sqrt{1-\rho} \\ \sqrt{1-\rho} & -\sqrt{\rho} \end{pmatrix}, \rho = f(p_t(i,h)) = f\left(\underset{j\in\Gamma_t(i)}{max}\left(p_t(i,j)\right)\right). \tag{4}
$$

It can be easily verified that the transformation $C$ is unitary, which satisfies the requirement of reversibility in Quantum Mechanics.

As is known, each term in a superposition state may be viewed as a position that the particle is located at and indicates the probability that it is found at that place. If the transformation $U$ is applied to the initial state twice or more, then more terms are yielded in the superposition state $|\psi\rangle$, which provides more positions to appear for the particle. Furthermore, these very positions enlarge the search area in the solution space and supply opportunities to obtain better results. Thanks to the quantum parallelism [1], it needs only one unitary operation to compute all positions and the corresponding probabilities appearing on them, which is unconceivable in a classical world.

Further, in the 1D-scms algorithm the total transformation $U = S \cdot (C \otimes I)$ will be applied to the initial state continuously $r$ times, so the step lengths moving left and right,$l_L$ and $l_R$, are reduced by a factor of $r$, namely $l_L/r$, $l_R/r$. Thus, using a method similar to Eq. (2), a conditional shift transformation $S$ can be expressed as the following unitary operator.

$$
S = |\uparrow\rangle\langle\uparrow| \otimes \sum_b |b + l_R/r\rangle\langle b| + |\downarrow\rangle\langle\downarrow| \otimes \sum_b |b - l_L/r\rangle\langle b| \tag{5}
$$

If the initial state of a particle is $|\psi_0\rangle = |\uparrow\rangle \otimes |0\rangle$, after the total transformation is applied $r = 2$ times, the obtained superposition state $|\psi\rangle$ is given as

following

$$
\begin{aligned}
|\psi_0\rangle &\xrightarrow{U} \sqrt{\rho}\,|\uparrow\rangle \otimes |l_R/r\rangle + \sqrt{1-\rho}\,|\downarrow\rangle \otimes |-l_L/r\rangle \\
&\xrightarrow{U} \rho\,|\uparrow\rangle \otimes |l_R\rangle + \sqrt{\rho(1-\rho)}\,|\downarrow\rangle \otimes |(l_R-l_L)/r\rangle \\
&\quad +(1-\rho)\,|\uparrow\rangle \otimes |(l_R-l_L)/r\rangle - \sqrt{\rho(1-\rho)}\,|\downarrow\rangle \otimes |-l_L\rangle = |\psi\rangle.
\end{aligned} \tag{6}
$$

From Eq. (6), we can see that the particle is not only on the position $l_R$ and $l_L$ with probabilities $\rho^2$ and $\rho(1-\rho)$, but also at the same time appears on a new position $(l_R - l_L)/r$ with probability $(1-\rho)$. At this time, if the superposition state is measured, it will collapse to one of three positions with probability. The corresponding component of the $m$-dimensional vector $\boldsymbol{X}_t^i$ will be updated according to the following formulation,

$$
\begin{aligned}
&\boldsymbol{X}_{t+1}^i(j) = \begin{cases} \boldsymbol{X}_t^i(j) + l_R & \text{if on position } l_R \\ \boldsymbol{X}_t^i(j) + (l_R - l_L)/r & \text{if on position } (l_R - l_L)/r \\ \boldsymbol{X}_t^i(j) - l_L & \text{if on position } -l_L \end{cases} \\
&l_R = \rho \times (\boldsymbol{X}_t^h(j) - \boldsymbol{X}_t^i(j)), j \in \{1, 2, \cdots, m\} \\
&l_L = (1-\rho) \times (\boldsymbol{X}_t^h(j) - \boldsymbol{X}_t^i(j)), j \in \{1, 2, \cdots, m\}.
\end{aligned} \tag{7}
$$

As data points move in space at random, their positions are constantly changing and the nearest neighbors of each data point also vary over time. So, the distances among data points and the degree of each data point need to be recomputed in the process. When the whole process is repeated until the sum of walking step lengths of all data points is less than a preset threshold $\varepsilon$, the algorithm exits.

## 3.2 1D-multi-coin-multi-step (1D-mcms) algorithm

In the previous algorithm, only the largest transition probability is used to establish the transformation $C$. If every one of all transition probabilities $p_t(i,j), j \in \Gamma_t(i)$ is employed to construct a transformation $C_j$ and different step lengths are used, $|\Gamma_t(i)|$ total transformations, $U_j = S_j \cdot (C_j \otimes I), j = 1, 2, \cdots, |\Gamma_t(i)|$, will be produced, where the symbol $|\cdot|$ represents the cardinality of a set. When all transformations $U_j$ are applied respectively, the probability distribution on the positions will be largely different from that in the 1D-scms algorithm.

Therefore, after the transition probabilities $p_t(i,j), j \in \Gamma_t(i)$ are computed, $k = |\Gamma_t(i)|$ transformations $C_j$ and conditional shift transformations $S_j$ are established. So

$$
\begin{aligned}
&U_j = S_j \cdot (C_j \otimes I) \\
&= \left(|\uparrow\rangle\langle\uparrow| \otimes \textstyle\sum_b |b + l_{R,j}\rangle\langle b| + |\downarrow\rangle\langle\downarrow| \otimes \textstyle\sum_b |b - l_{L,j}\rangle\langle b|\right) \\
&\cdot \left( \begin{pmatrix} \sqrt{\eta_j} & \sqrt{1-\eta_j} \\ \sqrt{1-\eta_j} & -\sqrt{\eta_j} \end{pmatrix} \otimes I \right), \eta_j = f(p_t(i,j)), j \in \Gamma_t(i).
\end{aligned} \tag{8}
$$

When all transformations $U_j$ are applied to the initial state $|\psi_0\rangle$, the obtained superposition state is $|\psi\rangle = U_k \cdots U_2(U_1|\psi_0\rangle)$.

If the initial state of a particle is $|\psi_0\rangle = |\uparrow\rangle \otimes |0\rangle$ and $k = 2$, after the transformations $U_1, U_2$ are applied, the superposition state $|\psi\rangle$ takes the form as below

$$
\begin{aligned}
|\psi_0\rangle &\xrightarrow{U} \sqrt{\eta_1}\,|\uparrow\rangle \otimes |l_{R,1}/k\rangle + \sqrt{1-\eta_1}\,|\downarrow\rangle \otimes |-l_{L,1}/k\rangle \\
&\xrightarrow{U} \sqrt{\eta_1\eta_2}\,|\uparrow\rangle \otimes |(l_{R,1}+l_{R,2})/k\rangle + \sqrt{\eta_1(1-\eta_2)}\,|\downarrow\rangle \otimes |(l_{R,1}-l_{L,2})/k\rangle \\
&+\sqrt{(1-\eta_1)(1-\eta_2)}\,|\uparrow\rangle \otimes |(l_{R,2}-l_{L,1})/k\rangle - \sqrt{\eta_2(1-\eta_1)}\,|\downarrow\rangle \otimes |-(l_{L,1}+l_{L,2})/k\rangle = |\psi\rangle.
\end{aligned}
\tag{9}
$$

From Eq. (9), we can see that the particle appears on four positions with different probabilities. Similarly, when the superposition state is measured, it will collapse to one of four positions with probability. The corresponding component of the $m$-dimensional vector $\boldsymbol{X}_t^i$ will be updated by the following formulation,

$$
\begin{aligned}
&\boldsymbol{X}_{t+1}^i(j) = \begin{cases} \boldsymbol{X}_t^i(j) + (l_{R,1}+l_{R,2})/k, & \text{if on position } (l_{R,1}+l_{R,2})/k \\ \boldsymbol{X}_t^i(j) + (l_{R,1}-l_{L,2})/k, & \text{if on position } (l_{R,1}-l_{L,2})/k \\ \boldsymbol{X}_t^i(j) + (l_{R,2}-l_{L,1})/k, & \text{if on position } (l_{R,2}-l_{L,1})/k \\ \boldsymbol{X}_t^i(j) - (l_{L,1}+l_{L,2})/k & \text{if on position } -(l_{L,1}+l_{L,2})/k \end{cases} \\
&l_{R,1} = \eta_1 \times (\boldsymbol{X}_t^h(j) - \boldsymbol{X}_t^i(j)),\ l_{L,1} = (1-\eta_1) \times (\boldsymbol{X}_t^h(j) - \boldsymbol{X}_t^i(j)), \\
&l_{R,2} = \eta_2 \times (\boldsymbol{X}_t^h(j) - \boldsymbol{X}_t^i(j)),\ l_{L,2} = (1-\eta_2) \times (\boldsymbol{X}_t^h(j) - \boldsymbol{X}_t^i(j)), \\
&j \in \{1, 2, \cdots, m\}.
\end{aligned}
\tag{10}
$$

The steps of four algorithms are summarized in Table 1.

# 4 Discussion

In the section, firstly, we discuss how the number of clusters is affected by the number $k$ of nearest neighbors changing. Then, in the 1D-scms and 1D-mcms algorithms, the relationship between the steps (times applying the total transformation) and the clustering accuracies of algorithms are investigated.

## 4.1 Number of nearest neighbors vs. number of clusters

The number $k$ of nearest neighbors represents the number of neighbors to which a data point $\boldsymbol{X}_t^i \in \boldsymbol{X}$ connects. If the longest distance among the data point and its $k$ nearest neighbors is selected as a radius, then a virtual circle centered around the data point can be drawn. This circle may be viewed as the interaction range of the data point whose radius follows the increase of the number of nearest neighbors. For a dataset, the number $k$ of nearest neighbors determines the number of clusters in part. Generally speaking, the number of clusters decreases with the increase of the number of nearest neighbors. For example, if the number $k$ of nearest neighbors is small, the interaction range of a data point is small too. Further, considering connectivity of a graph, we can find that the interaction ranges of data points intersect each other slightly, so that the connected domain formed is also small. In the process of data points moving in space, they will be close to one another gradually, which causes both the interaction ranges of data points and the connected domain on a graph are decreased further. Hence, in this case, they gather together only with not-too-distance data points around them. As a consequence, all data points form many small clusters, as is shown in Fig. 3(a).

Table 1: Steps of clustering algorithm.

| |
|---|
| Select a distance function $d(\cdot,\cdot)$ |
| Initialization: |
| Set the number of nearest neighbors $k$ and the separating threshold $\theta$ |
| Compute initial distance matrix $D(0)_{n\times n} = [d(\boldsymbol{X}_0^i, \boldsymbol{X}_0^j)]_{i,j=1,2,\cdots,n}$ |
| and initial degree vector $Deg(0)_{n\times 1} = [Deg_0(i)]_{i=1,2,\cdots,n}$ |
| Repeat: |
| Compute current distance matrix $D(t)_{n\times n} = [d(\boldsymbol{X}_t^i, \boldsymbol{X}_t^j)]_{i,j=1,2,\cdots,n}$ |
| Identify the current neighbor set $\Gamma_t(i)$ for each data point |
| Compute the current degree vector $Deg(t)_{n\times 1} = [Deg_t(i)]_{i=1,2,\cdots,n}$ |
| For each data point $\boldsymbol{X}_t^i$ |
| Compute transition probabilities $p_t(i,j), j \in \Gamma_t(i)$ according to Eq. 3 |
| 1D-scms: Establish the coin transformation $C$ and |
| the transformation $U$ using Eq. 4 and Eq. 8 |
| Apply the transformation to the initial state $t$ times in each dimension |
| Measure the state and update each component of $\boldsymbol{X}_t^i$ according to Eq. 10 |
| 1D-mcms: Establish the transformation $U_j$ using Eq. 11 |
| Apply the transformation to the initial state $k$ times in each dimension |
| Measure the state and update each component of $\boldsymbol{X}_t^i$ according to Eq. 13 |
| Compute sum of transition distances of each data point $\omega_i = \sum_{j=1}^m l_j$ |
| End For |
| Until $\sum_{i=1}^n \omega_i < \varepsilon$ |

On the other hand, if the number $k$ of nearest neighbors is large, the interaction ranges of data points will be increased at the same time, which makes them intersect each other largely. Thus, the larger connected domains on a graph are formed. Even if the interaction ranges of data points are decreased due to data points moving and approaching each other in space, the larger connected domains are established in contrast to that when selecting a small number $k$ of nearest neighbors. Finally, several big clusters are formed. Fig. 3 illustrates the relationship between the number of clusters and the number of nearest neighbors. As is analyzed above, eight clusters are obtained by the clustering algorithm, when $k = 8$. As the number $k$ of nearest neighbors rises, five clusters are obtained when $k = 14$ in Fig. 3(b), three clusters when $k = 22$ in Fig. 3(c). So, if the exact number of clusters is not known in advance, different number of clusters may be achieved by adjusting the number $k$ of nearest neighbors in practice.

## 4.2 Effect of number of steps in the 1D-scms and 1D-mcms algorithms

In 1D-scms algorithm, the coin transformation $C$ is constructed on the basis of the largest transition probability, but the steps moving left and right are reduced to $l_L/r$ and $l_R/r$. If the total transformation $U = S \cdot (C \otimes I)$ is applied to the initial state $|\psi_0\rangle$ $r$ times, the probability distribution on the positions of the particle will be different from the distribution of using only one time. As the times applying the transformation $U$ grow, the probability appearing on the position $l_L$ or $l_R$ will drop constantly, while the probability locating at a position between $l_L$ and $l_R$ will rise, and later the position with the largest probability will also move and approach to $l_L$ or $l_R$. Hence, if this is carried to the extreme, in the limit case, the 1D-scms algorithm will approach to the distribution produced by applying the total transformation $U$ only once, but the the largest probability drops. For instance, in each dimension, assume the initial state of a particle is $|\psi_0\rangle = |\uparrow\rangle \otimes |0\rangle$; the transition probabilities moving right and left are $\rho$ and $1 - \rho(\rho \geq 1 - \rho)$; and the steps are $l_L/r = l_R/r = 1$. After the transformation $U$ is applied to the initial state $t = 30, 60, 100$ times respectively, when $\rho = 0.8$, the probability distribution on the positions of the particle is illustrated in Fig. 4(a).

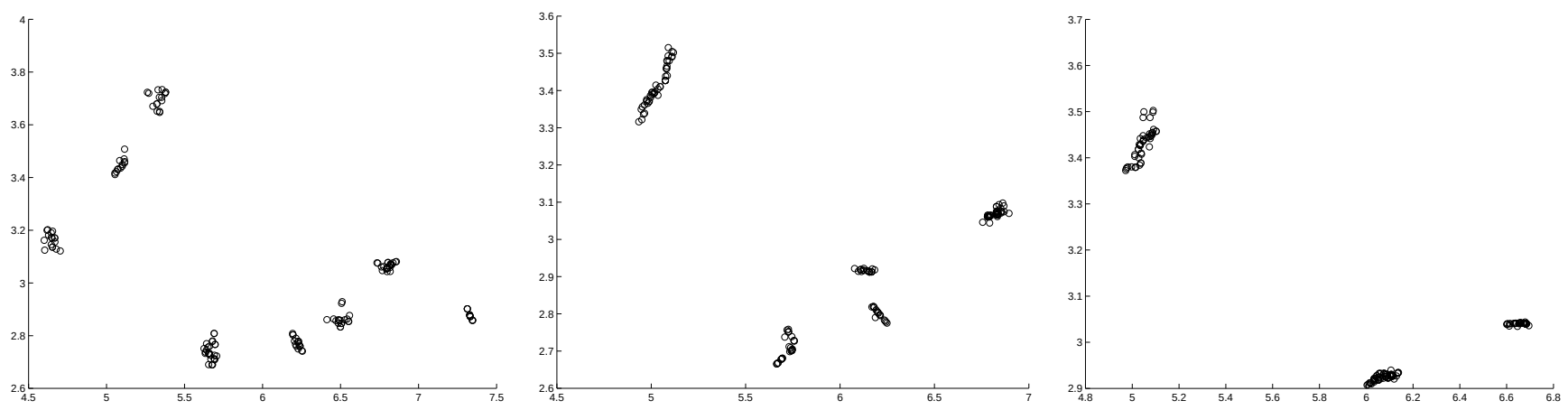

(a) $k = 8$ (b) $k = 14$ (c) $k = 22$

Figure 3: The number of nearest neighbors vs. number of clusters.

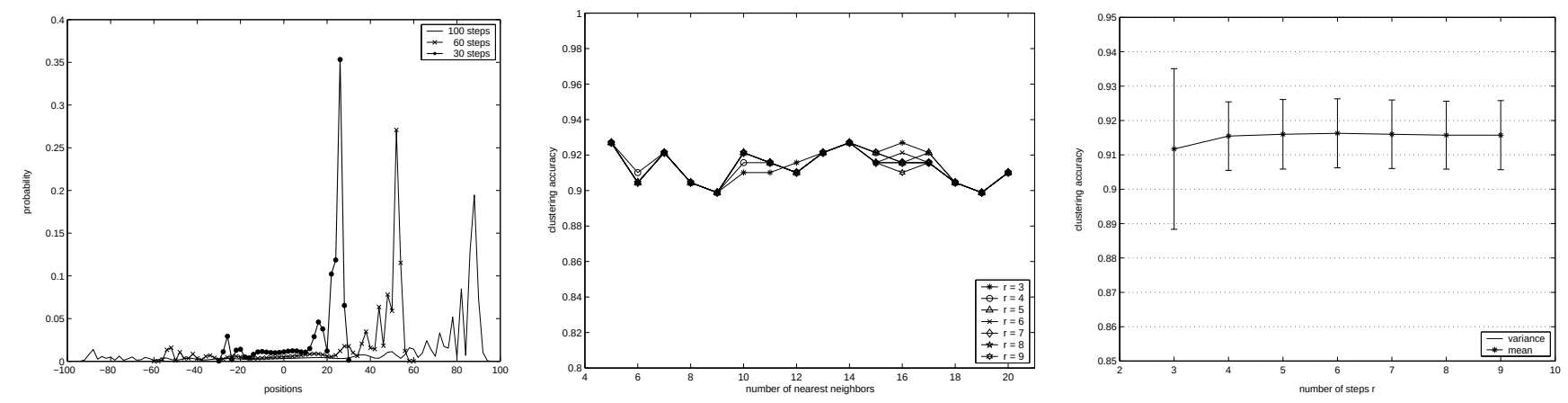

(a) (b) (c)

Figure 4: The steps vs. clustering accuracy in the 1D-scms algorithm.

As is shown in Fig. 4(a), for the same initial state, the probability distributions are different from each other, and the positions with the maximal probabilities tend to be close to the position $l_R$ due to the increase of the times $r$. Fig. 4(b) exhibits the relationship between the clustering accuracy (for definition, see the section 5.2) and the number of nearest neighbors, in which each curve is obtained at a fixed $r$. Further, the mean and variance of points in each curve is drawn in Fig. 4(c) which shows that the means are not monotonously increasing when the times vary, but drop slightly after $r = 6$, since too many times cause the degradation of the algorithm. So, for avoiding this, we recommend the times take $r = 5$ or $r = 6$.

The 1D-mcms algorithm differs from 1D-scms algorithm in that $k = |\Gamma_t(i)|$ transformations $C_j$ and $U_j = S_j \cdot (C_j \otimes I), j = 1, 2, \cdots, k$, are constructed and applied. As such, even if the same initial state $|\psi_0\rangle = |\uparrow\rangle \otimes |0\rangle$ is em-ployed, the obtained probability distribution on the positions is also different, which is shown in Fig. 5(a). Again, the number of total transformations is associated with the number of nearest neighbors directly. Thus, if the number of nearest neighbors is large, then more total transformations $U_j$ will be established, and the times applying them increases naturally. Similarly, from Fig. 5(c), we can observe that the degradation occurs in 1D-mcms algorithm too, when the times are large. Therefore, in practice, $k$ transition probabilities $p_t(i,j), j \in \Gamma_t(i)$ are sorted in descending order firstly, and then the first $r$ largest transition probabilities are selected to construct the transformations $U_j = S_j \cdot (C_j \otimes I), j = 1, 2, \cdots, r < k$, which is a method to reduce the degradation of the algorithm. Likewise, we recommend the times take $r = 5$ or $r = 6$ in the 1D-mcms algorithm.

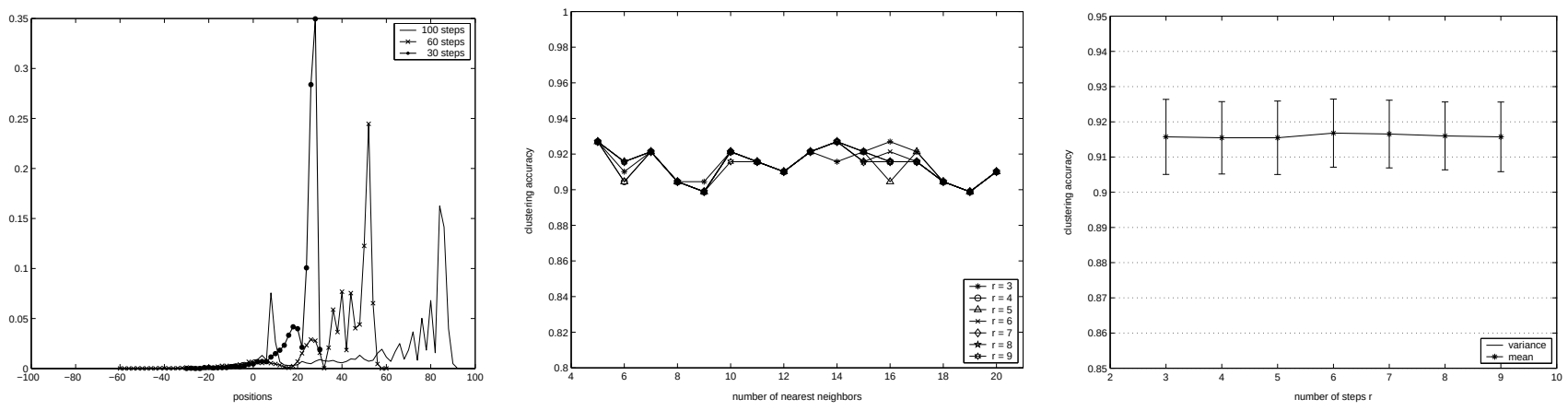

(a) (b) (c)

Figure 5: The steps vs. clustering accuracy in the 1D-mcms algorithm.

# 5 Experiment

To evaluate these four clustering algorithms, we choose six datasets from UCI repository [16], which are Soybean, Iris, Sonar, Glass, Ionosphere and Breast cancer Wisconsin datasets, and complete the experiments on them. In this section, firstly we introduce these datasets briefly, and then demonstrate the experimental results.

## 5.1 Experimental setup

The original data points in above datasets all are scattered in high dimensional spaces spanned by their features, where the description of all datasets is summarized in Table 2. As for Breast dataset, those lost features are replaced by random numbers. Finally, this algorithm is coded in Matlab 6.5.

Table 2: Description of datasets.

| Dataset | Instances | Features | Classes |
|---|---|---|---|
| Soybean | 47 | 21 | 4 |
| Iris | 150 | 4 | 3 |
| Sonar | 208 | 60 | 2 |
| Glass | 214 | 9 | 6 |
| Ionosphere | 351 | 32 | 2 |
| Breast | 699 | 9 | 2 |

Throughout all experiments, data points in a dataset are considered as movable particles whose initial positions are taken from the datasets directly. Next, the $k$ nearest neighbors of each data point in the dataset may be found, after a distance function is selected which only needs to satisfy that the more similar data points are, the smaller the output of the function is. In the experiments, the Euclidean distance function, $L2$-norm distance, is employed. Additionally, in the 1D-scms and 1D-mcms algorithms, the variable $r$ is set at $r = 6$.

### 5.2 Experimental results

Two above-constructed clustering algorithms are experimented on the six datasets respectively. As is analyzed in section 4.1, for a dataset the number of clusters decreases with the increase of the number $k$ of nearest neighbors. Therefore, when a small $k$ is selected, it is possible that the number of clusters is larger than the preset number of clusters in the dataset, after the algorithm is end. So a merging-subroutine is called to merge unwanted clusters, which works in this way. At first, the cluster with the fewest data points is identified, and then is merged to the cluster whose distance between their centroids is smallest. This subroutine is repeated till the number of clusters is equal to the preset number. Moreover, the algorithms are run on every dataset at the different number of nearest neighbors, and clustering results obtained by these four algorithms are compared in Fig. 6, in which each point represents a clustering accuracy.

**Definition 1** *$cluster_i$ is the label which is assigned to a data point $\boldsymbol{X}_i$ in a dataset by the algorithm, and $c_i$ is the actual label of the data point $\boldsymbol{X}_i$ in the dataset. So the clustering accuracy is [17]:*

$$\begin{aligned} accuracy &= \frac{\sum_{i=1}^{N} \lambda\Big(map(cluster_i), c_i\Big)}{N} \\ \lambda(map(cluster_i), c_i) &= \begin{cases} 1 & if\ map(cluster_i) = c_i \\ 0 & otherwise \end{cases} \end{aligned} \tag{11}$$

*where the mapping function $map(\cdot)$ maps the label got by the algorithm to the actual label.*

As is shown in Fig. 6, the similar results are obtained by these algorithms at different nearest neighbors, but almost all the best results are yielded by the 1D-mcms algorithm. Additionally, we compare our results to those results obtained by other clustering algorithms, Kmeans [18], PCA-Kmeans [18], LDA-Km [18], on the same dataset. The comparison is summarized in Table 3.

Table 3: Comparison of clustering accuracies of algorithm.

| Algorithm | Soybean | Iris | Sonar | Glass | Ionosphere | Breast |
|---|---|---|---|---|---|---|
| 1D-scms | 91.49% | 90% | 62.02% | 64.49% | 71.51% | 95.42% |
| 1D-mcms | 97.87% | 96.67% | 62.02% | 64.02% | 75.21% | 95.42% |
| Kmeans | 68.1% | 89.3% | – | 47.2% | 71% | – |
| PCA-Kmeans | 72.3% | 88.7% | – | 45.3% | 71% | – |
| LDA-Km | 76.6% | 98% | – | 51% | 71.2% | – |

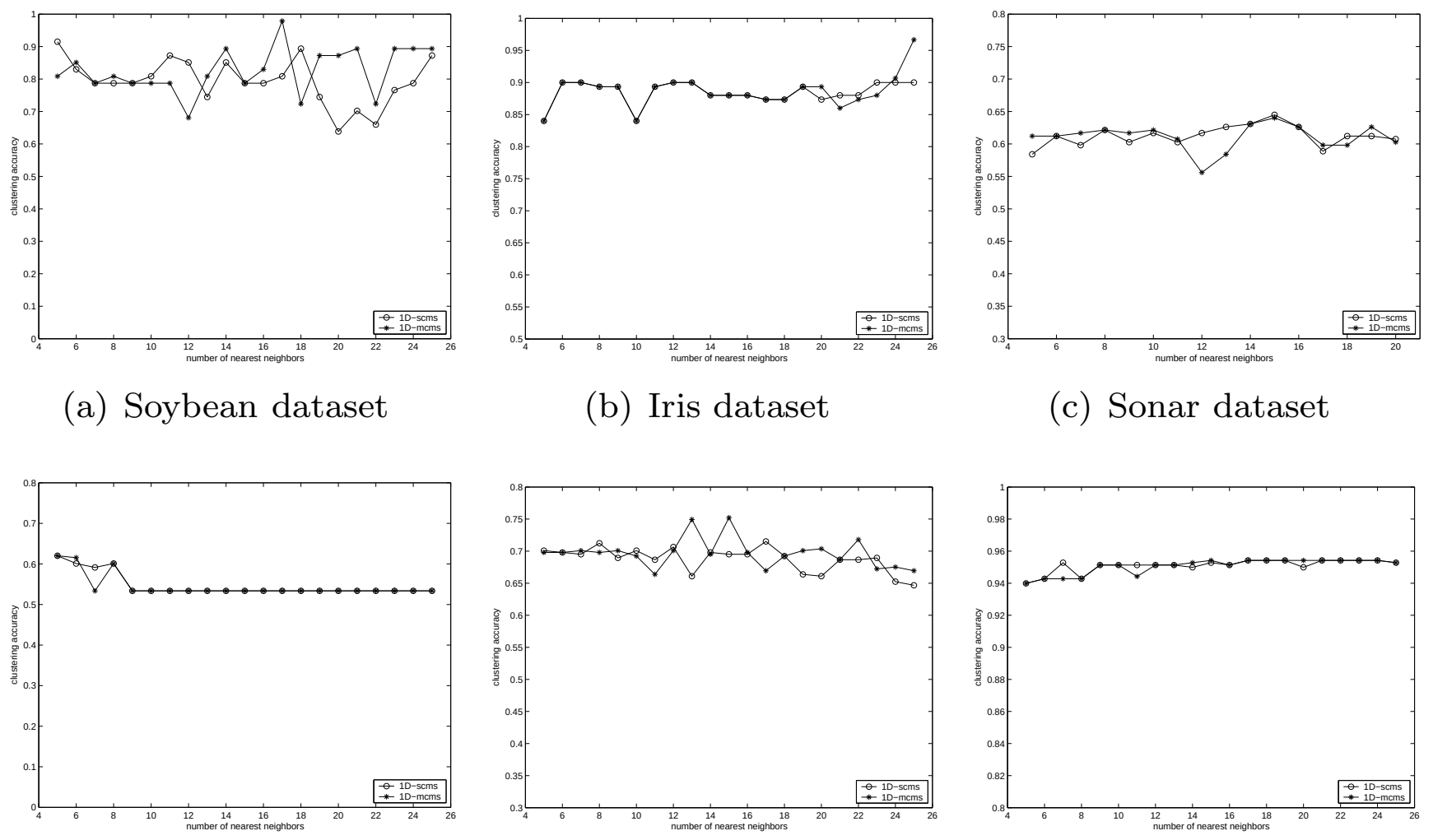

(a) Soybean dataset (b) Iris dataset (c) Sonar dataset

(d) Glass dataset (e) Ionosphere dataset (f) Breast dataset

Figure 6: Comparison of clustering accuracies in all proposed algorithms.

# 6 Conclusions

The enormous successes gained by the quantum algorithms make us realize it is possible that the quantum algorithms can obtain solutions faster and better than those classical counterparts. Therefore, we combine the QRW with the problem of data clustering, and establish four clustering algorithms based on the QRW. In the algorithms, those data points for clustering are considered as movable particles which may be also seen as local control subsystems from the point of view of control theory. Further, we develop two clustering algorithms based on the one dimensional QRW, and discuss the probability distributions on the positions induced by the QRW under two different cases: (a) only one transformation $C$ is constructed but the total transformation $U$ are applied twice or more; (b) more transformations $C$ and $U$ are constructed and applied. Thanks to the quantum parallelism, all positions and the corresponding probabilities are computed by applying the unitary operation only once in contrast to many times in classical world. Besides, on the basis of the QRW, the probability distributions on the positions that dose not exist in the classical case are produced, which provides opportunities for obtaining better results.

In these algorithms, when the exact number of clusters is unknown in advance, one can adjust the number $k$ of nearest neighbors to control the number of clusters which decreases with the increase of the number $k$ of nearest neighbors. We evaluate the clustering algorithms on six real datasets, and experimental results have demonstrated that data points in a dataset are clustered reasonably and efficiently. Additionally, these clustering algorithms can also detect clusters of arbitrary shape, size and density.

## Acknowledgments

This work is supported in part by the National Natural Science Foundation of China (No. 60405012, No. 60675055).

## References

[1] M. A. Nielsen and I. L. Chuang, *Quantum Computation and Quantum Information.* Cambridge: Cambridge University Press, 2000.

[2] P. W. Shor, "Algorithms for quantum computation: discrete logarithms and factoring," in *Foundations of Computer Science, 1994 Proceedings., 35th Annual Symposium on*, pp. 124–134, 1994.

[3] L. K. Grover, "Quantum mechanics helps in searching for a needle in a haystack," *Physical Review Letters*, vol. 79, no. 2, p. 325, 1997.

[4] G. Brassard, P. Høyer, and A. Tapp, "Quantum counting," in *Automata, Languages and Programming*, pp. 820–831, 1998.

[5] A. Ambainis, E. Bach, A. Nayak, A. Vishwanath, and J. Watrous, "One-dimensional quantum walks," in *Proceedings of the thirty-third annual ACM symposium on Theory of computing*, (Hersonissos, Greece), ACM, 2001.

[6] T. A. Brun, H. A. Carteret, and A. Ambainis, "Quantum walks driven by many coins," *Physical Review A*, vol. 67, no. 5, p. 052317, 2003.

[7] J. Kempe, "Quantum random walks: an introductory overview," *Contemporary Physics*, vol. 44, no. 4, pp. 307 – 327, 2003.

[8] D. Aharonov, A. Ambainis, J. Kempe, and U. Vazirani, "Quantum walks on graphs," in *Proceedings of the thirty-third annual ACM symposium on Theory of computing*, (Hersonissos, Greece), pp. 50–59, ACM, 2001.

[9] J. Kempe, "Discrete quantum walks hit exponentially faster," *Probability Theory and Related Fields*, vol. 133, no. 2, pp. 215–235, 2005.

[10] A. M. Childs, R. Cleve, E. Deotto, E. Farhi, S. Gutmann, and D. A. Spielman, "Exponential algorithmic speedup by a quantum walk," in *Proceedings of the thirty-fifth annual ACM symposium on Theory of computing*, (San Diego, CA, USA), ACM, 2003.

[11] E. Aïmeur, G. Brassard, and S. Gambs, "Quantum clustering algorithms," in *Proceedings of the 24th International Conference on Machine Learning*, (Corvallis, OR), 2007.

[12] D. Horn and A. Gottlieb, "Algorithm for data clustering in pattern recognition problems based on quantum mechanics," *Physical Review Letters*, vol. 88, no. 1, p. 018702, 2001.

[13] R. Schützhold, "Pattern recognition on a quantum computer," *Physical Review A*, vol. 67, no. 6, p. 062311, 2003.

[14] A. Ekert, P. M. Hayden, and H. Inamori, "Course 10: Basic concepts in quantum computation," in *Coherent atomic matter waves*, vol. 72, p. 661, Springer Berlin / Heidelberg, 2001.

[15] Y. Omar, N. Paunković, L. Sheridan, and S. Bose, "Quantum walk on a line with two entangled particles," *Physical Review A (Atomic, Molecular, and Optical Physics)*, vol. 74, no. 4, pp. 042304–7, 2006.

[16] C. Blake and C. Merz, *UCI Repository of machine learning databases.* http://www.ics.uci.edu/mlearn/MLRepository.html: Department of ICS, University of California, Irvine., 1998.

[17] G. Erkan, "Language model-based document clustering using random walks," in *Proceedings of the main conference on Human Language Technology Conference of the North American Chapter of the Association of Computational Linguistics*, (New York, New York), Association for Computational Linguistics, 2006.

[18] C. Ding and T. Li, "Adaptive dimension reduction using discriminant analysis and k-means clustering," in *Proceedings of the 24th International Conference on Machine Learning*, (Corvallis, OR), pp. 521–528, 2007.